\documentclass[letterpaper, 10 pt, conference]{packages/ieeeconf}
\IEEEoverridecommandlockouts

\usepackage{packages/algorithm}
\usepackage{algorithmicx,algpseudocode}

\usepackage[compress]{cite}
\makeatletter
\let\NAT@parse\undefined
\makeatother

\usepackage[table]{xcolor} %
\usepackage[section]{placeins} %

\usepackage{graphics} %
\usepackage{graphicx}
\usepackage{amsmath} %
\usepackage{mathtools} %
\usepackage{todonotes}
\usepackage{amssymb}  %
\usepackage{makeidx} %
\usepackage{graphicx} %
\graphicspath{{figures/}}
\usepackage{multicol} %
\usepackage[bottom]{footmisc} %
\usepackage[utf8]{inputenc} %
\usepackage{multirow}
\usepackage{capt-of}
\usepackage{booktabs} %
\usepackage{adjustbox}
\usepackage[font=small]{caption}
\usepackage{makecell} %
\usepackage{subcaption} %
\usepackage{booktabs} %
\usepackage{bm} %
\usepackage{float} %
\usepackage{color} %
\usepackage{empheq} %
\usepackage{threeparttable} %
\usepackage{xcolor}
\usepackage{tcolorbox}
\usepackage{pifont}   %
\newif\ifenablepdfhyperlinks
\enablepdfhyperlinksfalse
\ifenablepdfhyperlinks
    \usepackage{hyperref}
\else
    \usepackage[draft]{hyperref}
\fi
\usepackage{todonotes}

\usepackage{amsthm}
\DeclareMathSymbol{\shortminus}{\mathbin}{AMSa}{"39}
\usepackage{paralist}
\newtheorem{thm}{Theorem}

\usepackage{mathtools}
\DeclarePairedDelimiterX{\norm}[1]{\lVert}{\rVert}{#1}

\allowdisplaybreaks
\usepackage{acronym}
\usepackage[acronym]{glossaries}

\acrodef{COM}[COM]{Center Of Mass}
\acrodef{LIP}[LIP]{Linear Inverted Pendulum}
\acrodef{SLIP}[SLIP]{Spring Loaded Inverted Pendulum}
\acrodef{SRB}[SRB]{Single Rigid Body}
\acrodef{HZD}[HZD]{Hybrid Zero Dynamics}
\acrodef{MPC}[MPC]{Model Predictive Control}
\acrodef{ZMP}[ZMP]{Zero Moment Point}
\acrodef{DDP}[DDP]{Differential Dynamic Programming}
\acrodef{MIQCQP}[MIQCQP]{Mixed-Integer Quadratically-Constrained Quadratic Program}
\acrodef{DRL}[DRL]{Deep Reinforcement Learning}
\acrodef{PPO}[PPO]{Proximal Policy Optimization}
\acrodef{FNN}[FNN]{Feedforward Neural Network}
\acrodef{LSTM}[LSTM]{Long Short-Term Memory}
\acrodef{DMPO}[DMPO]{Distributional Maximum a posteriori Policy Optimization}
\acrodef{SLAM}[SLAM]{simultaneous localization and mapping}
\acrodef{GTSAM}[\textsc{gtsam}]{GATech Smoothing And Mapping}
\acrodef{EKF}[EKF]{extended Kalman filter}
\acrodef{iSAM}[i\textsc{sam}]{incremental smoothing and mapping}
\acrodef{RNN}[\textsc{rnn}]{recurrent neural networks}
\acrodef{EM}[\textsc{em}]{expectation maximization}
\acrodef{PINN}[\textsc{pinn}]{physics-informed neural networks}
\acrodef{FoV}[\textsc{f}o\textsc{v}]{field of view}
\acrodef{PDE}[\textsc{pde}]{partial differential equation}
\acrodef{DSS}[\textsc{dss}]{decision support system}
\acrodef{SARA}[\textsc{sara}]{Scalable, Adaptive, and Resilient Autonomy}
\acrodef{MSG-Cal}[\textsc{msg-c}al]{multi-sensor graph-based calibration}
\acrodef{IMU}[IMU]{inertial measurement unit}
\acrodef{VINS}[\textsc{vins}]{visual inertial navigation system}
\acrodef{KF}[\textsc{kf}]{Kalman filter}
\acrodef{MSCKF}[MSCKF]{multi-state constraint Kalman filter}
\acrodef{KLD}[\textsc{kld}]{Kullback-Leibler divergence}
\acrodef{RMSE}[RMSE]{root mean square error}
\acrodef{NEES}[\textsc{nees}]{normalized estimation error square}
\acrodef{INS}[\textsc{ins}]{inertial navigation system}
\acrodef{APG}[\textsc{apg}]{Aberdeen Proving Grounds}
\acrodef{FOA}[\textsc{foa}]{funding opportunity announcement}
\acrodef{UAV}[UAV]{unmanned aerial vehicle}
\acrodef{ASD}[\textsc{asd}]{ Autonomous Systems Division }
\acrodef{VIWO}[\textsc{viwo}]{visual-inertial-wheel odometry}
\acrodef{VIO}[\textsc{vio}]{visual-inertial odometry}
\acrodef{GPS}[\textsc{gps}]{global positioning system}
\acrodef{ATE}[\textsc{ate}]{absolute trajectory error}
\acrodef{ENU}[\textsc{enu}]{east-north-up}
\acrodef{CPU}[\textsc{cpu}]{central processing unit}
\acrodef{CRADA}[\textsc{crada}]{cooperative research and development agreement}
\acrodef{VTD}[\textsc{vtd}]{Vehicle Technology Directorate}
\acrodef{CL}[\textsc{cl}]{cooperative localization}
\acrodef{GPU}[\textsc{gpu}]{graphics processing unit}
\acrodef{CNN}[CNN]{convolutional neural network}
\acrodef{LSD-SLAM}[\textsc{lsd-slam}]{large scale direct \textsc{slam}}
\acrodef{ORB-SLAM}[\textsc{orb-slam}]{oriented \textsc{fast} and rotated \textsc{brief} \textsc{slam}}
\acrodef{VI-SLAM}[VI-SLAM]{visual-inertial \textsc{slam}}
\acrodef{RPE}[\textsc{rpe}]{relative pose error}
\acrodef{CA}[\textsc{ca}]{cooperative agreement}
\acrodef{RAS}[\textsc{ras}]{Robotic Autonomous Systems}
\acrodef{VST}[\textsc{vst}]{variable stiffness treadmill}
\acrodef{RAC}[\textsc{rac}]{Robotic Autonomous Systems}
\acrodef{COM}[COM]{center of mass}
\acrodef{QP}[QP]{quadratic programming}
\acrodef{FOV}[FOV]{field of view}
\acrodef{GS}[\textsc{gs}]{Gaussian splatting}
\acrodef{SDF}[\textsc{sdf}]{signed distance function}
\acrodef{MINS}[MINS]{multisensor-aided inertial navigation system}
\acrodef{GNSS}[\textsc{gnss}]{global navigation satellite system}
\acrodef{MSF}[MSF]{multi-sensor fusion}
\acrodef{DNN}[\textsc{dnn}]{deep neural network}
\acrodef{POC}[\textsc{poc}]{point of contact}
\acrodef{MAP}[m\textsc{AP}]{mean average precision}
\acrodef{ISR}[ISR]{intelligence, surveillance and reconnaissance}
\acrodef{FOA}[FOA]{funding opportunity announcement}
\acrodef{LIP}[LIP]{Linear Inverted Pendulum}
\acrodef{SLIP}[SLIP]{Spring Loaded Inverted Pendulum}
\acrodef{HZD}[HZD]{Hybrid Zero Dynamics}
\acrodef{MPC}[MPC]{model predictive control}
\acrodef{ZMP}[ZMP]{zero moment point}
\acrodef{DDP}[DDP]{differential dynamic programming}
\acrodef{MIQCQP}[MIQCQP]{mixed-integer quadratically-constrained quadratic program}
\acrodef{DRL}[DRL]{deep reinforcement learning}
\acrodef{PPO}[PPO]{proximal policy optimization}
\acrodef{FNN}[FNN]{feedforward neural network}
\acrodef{LSTM}[LSTM]{long short-term memory}
\acrodef{DMPO}[DMPO]{distributional maximum a-posteriori Policy Optimization}
\acrodef{UD}[UD]{University of Delaware}
\acrodef{DSU}[DSU]{Delaware State University}
\acrodef{MSI}[MSI]{minority serving institution}
\acrodef{OSF}[OSF]{Open Science Framework}
\acrodef{NeRF}[NeRF]{neural radiance field}
\acrodef{3DGS}[3DGS]{3D Gaussian splatting}
\acrodef{MVS}[MVS]{multi-view stereo}
\acrodef{OSC}[OSC]{operations support center}
\acrodef{STAR}[STAR]{science, technology and research}
\acrodef{DTRA}[DTRA]{Defense Threat Reduction Agency}
\acrodef{EHS}[EHS]{Office of Environmental Health \& Safety}
\acrodef{SfM}[SfM]{structure from motion}
\acrodef{COTS}[COTS]{commercial off-the-shelf}
\acrodef{ROS}[ROS]{Robot Operating System}
\acrodef{PSNR}[PSNR]{peak signal-to-noise ratio}
\acrodef{SSIM}[SSIM]{structural similarity index measure}
\acrodef{COM}[COM]{Center Of Mass}
\acrodef{ROM}[ROM]{Reduced-order Model}
\acrodef{LIP}[LIP]{Linear Inverted Pendulum}
\acrodef{SLIP}[SLIP]{Spring Loaded Inverted Pendulum}
\acrodef{HZD}[HZD]{Hybrid Zero Dynamics}
\acrodef{WBC}[WBC]{Whole Body Control}
\acrodef{RL}[RL]{Reinforcement Learning}
\acrodef{MPC}[MPC]{Model Predictive Control}
\acrodef{DMPC}[DMPC]{Deterministic Model Predictive Control}
\acrodef{SMPC}[SMPC]{Stochastic Model Predictive Control}
\acrodef{PSSMPC}[PSSMPC]{Probabilistically Scaling Stochastic Model Predictive Control}
\acrodef{ZMP}[ZMP]{Zero Moment Point}
\acrodef{COP}[COP]{Center of Pressure}
\acrodef{QP}[QP]{Quadratic Program}
\acrodef{CLF}[CLF]{Control Lyapunov Function}
\acrodef{CBF}[CBF]{Control Barrier Function}
\acrodef{DDP}[DDP]{Differential Dynamic Programming}
\acrodef{MIQCQP}[MIQCQP]{Mixed-Integer Quadratically-Constrained Quadratic Program}
\acrodef{DRL}[DRL]{Deep Reinforcement Learning}
\acrodef{PPO}[PPO]{Proximal Policy Optimization}
\acrodef{FNN}[FNN]{Feedforward Neural Network}
\acrodef{LSTM}[LSTM]{Long Short-Term Memory}
\acrodef{DMPO}[DMPO]{Distributional Maximum a posteriori Policy Optimization}
\acrodef{iid}[i.i.d.]{Independent and Identically Distributed}
\acrodef{iLQR}[iLQR]{Iterated LQR}
\acrodef{RNN}[RNN]{Recurrent Neural Network}
\acrodef{GRU}[GRU]{Gated Recurrent Unit}
\acrodef{ESDF}[ESDF]{Euclidean Signed Distance Field}
\acrodef{SDF}[SDF]{Signed Distance Field}
\acrodef{SAS}[SAS]{Simple Approximating Set}
\acrodef{MPPI}[MPPI]{Model Predictive Path Integral}
\acrodef{PAC}[PAC]{Probably Approximately Correct}
\acrodef{dEBF}[dEBF]{Discrete Exponential Barrier Function}
\acrodef{VLN}[VLN]{Vision-Language Navigation}
\acrodef{VLA}[VLA]{Vision-Language Action}
\acrodef{VLM}[VLM]{Vision-Language Model}
\acrodef{LLM}[LLM]{Large Language Model}
\acrodef{IL}[IL]{Imitation Learning}
\acrodef{CEM}[CEM]{Cross Entropy Method}

\usepackage{eso-pic}

\begin{document}
\title{\vspace{-2.5em}%
  \smash{\makebox[\textwidth][r]{\footnotesize\normalfont
    Accepted to IEEE-RAS Humanoids 2026
  }}\vspace{1.5em}\\
  \LARGE \bf Caging Vision-Language Navigation for Legged Robots \\via Barrier-Certified Safety Architecture
}

\author{Abhijeet Mangesh Kulkarni$^{1}$, Saimouli Katragadda$^{1}$, Yuze Du$^{1}$, Zijun Sha$^{2}$, and Guoquan Huang$^{1}$%
\thanks{This work was partially supported by 
the University of Delaware (UD) College of Engineering,  
Delaware NASA/EPSCoR Seed Grant, 
NSF (IIS-2410019), 
and Toyota Motor Engineering \& Manufacturing North America. 
}
\thanks{$^{1}$The authors are with the Robot Perception and Navigation Group (RPNG), University of Delaware, Newark, DE 19716, USA. Email: {\small\ttfamily \{amkulk, saimouli, yuzedu, ghuang\}@udel.edu}.}%
\thanks{$^{2}$Zijun Sha is with the Frontier Research Center, Toyota Motor Corporation, Tokyo, Japan. Email: {\small\ttfamily zijun\_sha@mail.toyota.co.jp}}%
}

\maketitle

\begin{abstract}
By harnessing the inherent commonsense capabilities of pretrained \acp{LLM} and \acp{VLM}, current \ac{VLA} models for legged robots excel at interpreting language-based navigation commands but lack the physical grounding and explicit safety assurance required for real-world operation.
Therefore, this paper aims to cage Legged-\ac{VLN} through a safety architecture that provides physical grounding and ensures navigation safety.
The key idea is not to abandon the powerful capabilities of end-to-end \ac{VLA} planning, but rather to augment and cage them.
In particular, we integrate a high-level \ac{VLA} planner with a low-level physical-grounding layer that leverages efficient state estimation, in-situ 3D spatial perception, and an \ac{SDF} representation to generate executable trajectories. We provide a navigation planner with provable collision-avoidance guarantees under the stated assumptions by combining a low-dimensional Bézier-parameterized MPPI planner with \ac{dEBF} constraints.
The proposed Legged-\ac{VLN} system is modular, making it compatible with various \ac{VLM}/\ac{VLA} models and agnostic to the robot platform.
The proposed system is validated with extensive real-world experiments on humanoid and quadrupedal robots, demonstrating versatile, efficient, and safe vision-language driven navigation capabilities.
\end{abstract}

\section{Introduction}
Legged robots, including humanoids and quadrupeds, are being engineered to work hand-in-hand with humans and perform practical, real-world tasks with precision and purpose (e.g. \cite{unitree2025a2_video}). A fundamental prerequisite for accomplishing such tasks is the ability to navigate safely to locations where subsequent actions can be performed. Traditionally, navigation missions are specified as sequences of goal poses through a navigation interface, which typically requires either a trained operator or an expert-authored behavior tree \cite{Macenski2020IROS}. Recent advances in \acp{VLM}/\acp{VLA}~\cite{Sapkota2025} have enabled \ac{VLN}, allowing users to specify navigation missions through intuitive natural-language instructions rather than manually defined waypoints. However, to deploy these robots alongside humans, it is imperative that \ac{VLN} systems are demonstrably safe—generating motion that are both semantically correct and collision free. This work directly addresses this critical challenge of ensuring physical safety in Legged-\ac{VLN}.

The core issue is that while pretrained models offer powerful visual-semantic reasoning, their navigation outputs~\cite{Cheng2025,Bu2025,Chen2025} typically ignore the robot's kinematics, local obstacle configurations, and formal safety constraints. Consequently, an end-to-end \ac{VLA} planner might propose an instruction-following navigation commands that are physically unexecutable or prone to collisions. Therefore, rather than executing these outputs directly, they must be treated as high-level navigation proposals that are physically grounded and verified. We refer to this verification process as \emph{caging}: the interposition of a safety-assured physical-grounding layer between the \ac{VLM}/\ac{VLA} planner and the robot's locomotion controller, preserving the reasoning capabilities of pretrained models while providing provable collision-avoidance guarantees under the stated assumptions.

To this end, we propose a dual-layer caging architecture for Legged-\ac{VLN} that integrates high-level semantic reasoning with low-level, safety-assured motion planning. We evaluate the complete framework in real-world indoor and outdoor environments using a Unitree G1 humanoid and a Ghost Robotics Vision 60 quadruped.

Specifically, the main contributions of this work are:
\begin{itemize}
    \item \textbf{A dual-layer caging architecture} that decouples semantic reasoning from physical execution by treating pretrained \ac{VLM}/\ac{VLA} outputs as high-level proposals and mapping them to locally feasible goal poses with open-vocabulary terminal alignment.
    \item \textbf{A safety-assured motion planner} utilizing an efficient, low-dimensional Bézier-parameterized \ac{MPPI} formulation with \ac{SDF}-based \ac{dEBF} constraints, providing provable collision-avoidance guarantees under stated assumptions.
    \item \textbf{Extensive real-world validation} on both humanoid and quadrupedal platforms, demonstrating cross-platform applicability and reliable, safe execution of natural-language navigation instructions.
\end{itemize}

\section{Related Work}

\begin{figure*}[t]
    \centering
    \includegraphics[width=.925\linewidth]{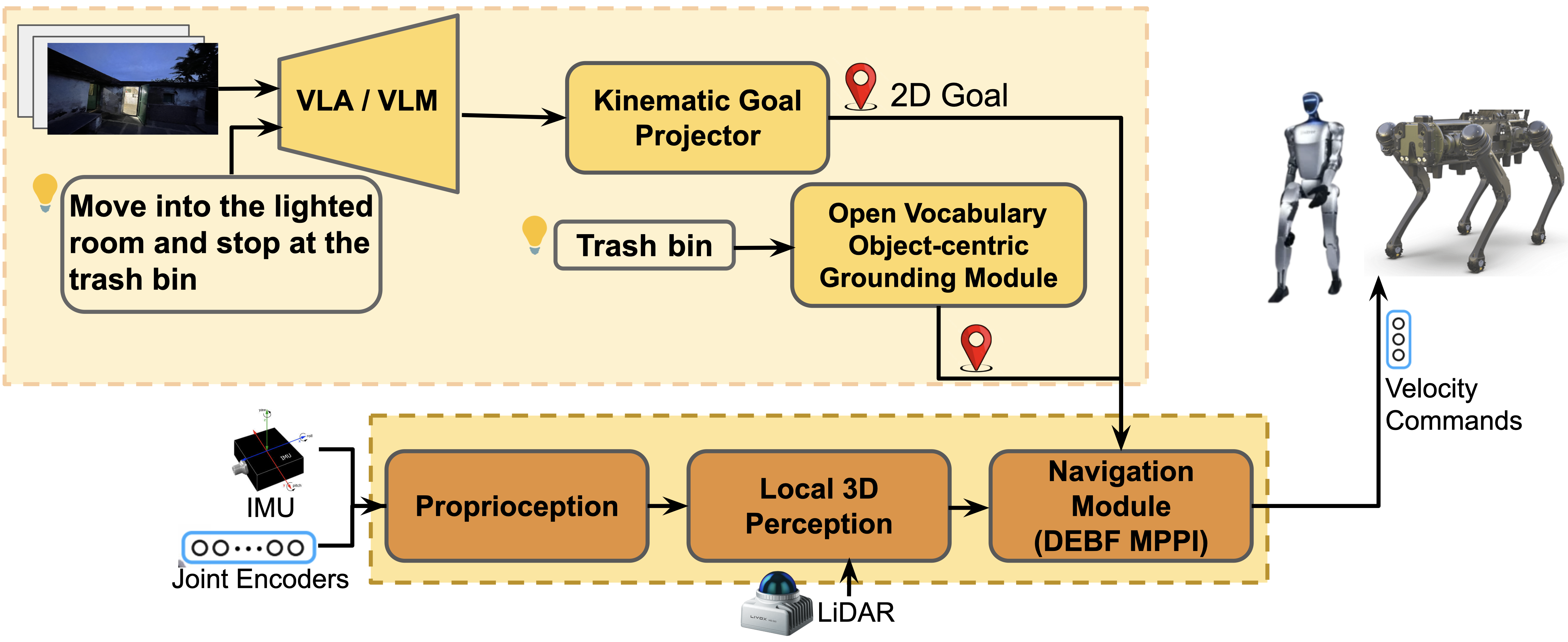}
    \caption{The proposed Legged-VLN architecture separates semantic grounding
    (top) from safety-assured physical grounding (bottom).}
    \label{fig:system_pipeline}
    \vspace{-10pt}
\end{figure*}

\noindent\textbf{Vision-language navigation for legged robots.}
VLN was formalized by the Room-to-Room benchmark~\cite{Anderson2018} and later
extended from discrete navigation graphs to continuous environments through
R2R-CE~\cite{Krantz2020}. Recent \acp{VLA} directly map visual and language
observations to robot actions. Of particular relevance, QUAR-VLA targets
quadruped control~\cite{Ding2024}, while NaVILA performs language-conditioned
navigation on legged robots~\cite{Cheng2025}. These models provide strong
semantic action proposals, but their learned outputs do not explicitly encode
the robot footprint, the current local obstacle geometry, or a formal
closed-loop safety condition. Recent safety-aware VLA methods address this gap
either by aligning the policy through constrained learning, as in
SafeVLA~\cite{Zhang2025SafeVLA}, or by correcting manipulation actions with a
plug-and-play \ac{CBF} layer, as in VLSA/AEGIS~\cite{Hu2025VLSA}. Our work instead
targets legged navigation: it leaves the semantic model unchanged and cages its
proposals using online geometric perception and horizon-level motion
certification. This separation allows the front end and robot platform to
change without transferring responsibility for navigation safety to the
learned model. \noindent\textbf{Semantic and geometric grounding.}
Methods such as 3D-LLM~\cite{Hong2023b} and Chat-Scene~\cite{Huang2024c} enrich
language models with 3D scene structure for spatial reasoning. Open-vocabulary
detection and segmentation models, including Grounding-DINO~\cite{Liu2023_arxiv}
and SAM2~\cite{Ravi2024_arxiv}, additionally enable
language-specified objects to be localized without a fixed task vocabulary.
Splat-Nav~\cite{Chen2025TRO} more tightly couples these domains by using a
language-embedded Gaussian-splat map to specify goals, construct safe
polytope corridors, and generate B\'{e}zier trajectories. Our architecture
instead uses the learned front end and open-vocabulary perception to generate
semantic goals and refine terminal alignment, while an incrementally updated
environmental \ac{SDF} grounds every proposed motion in locally observed free
space. Safety is therefore evaluated from geometric clearance rather than
semantic confidence. \noindent\textbf{Safety in sampling-based predictive control.}
For legged robots, sequential MPC over convex safe corridors enables reactive
bipedal planning in cluttered environments~\cite{Narkhede2022RAL}.
\ac{MPPI} supports derivative-free trajectory optimization with nonlinear and
nonsmooth costs~\cite{Williams2018_TRO}; low-dimensional trajectory
parameterizations further improve sampling efficiency in cluttered
environments~\cite{Kulkarni2026ICRA}. Discrete-time barrier conditions provide a
route to forward-invariant safety~\cite{Agrawal2017RSS,Zeng2021ACC}, and recent
sampling-based controllers incorporate barrier functions through shielded
refinement~\cite{Yin2023_arxiv}, barrier-rate guidance with learned
\acp{SDF}~\cite{Parwana2025_arxiv}, or discrete barrier
states~\cite{Wang2025_arxiv}. In contrast, our physical-grounding layer combines
B\'{e}zier-parameterized MPPI, value-only queries to an incrementally updated
environmental SDF, soft dEBF-guided sampling, and hard rollout certification
with a safe backup. This construction cages interchangeable VLM/VLA proposals
and is deployed unchanged at the planning level on both quadrupedal and
humanoid platforms.

\section{System Overview}
\label{sec:problem}

We address the problem of language-driven, safety-critical navigation for legged robots operating in cluttered, unknown environments. 
That is, given (i) a natural-language instruction describing a high-level task and optionally a target object, (ii) onboard sensing (e.g., vision, LiDAR, IMU, leg kinematics), and (iii) platform actuation limits, the objective is to synthesize a control sequence $\{\mathbf{u}_k\}_{k=0}^{T-1}$ that steers the robot from an initial pose $\mathbf{x}_0$ to a a goal region or an object-relative pose while being  caged to safely execute the commands.
Formally, the robot must: (i) ground the language into a sequence of high-level, spatially referenced subgoals; 
(ii) convert these into feasible, collision-free subgoals; and (iii) ensure safety at all times. 
The solution should be modular (compatible with different legged platforms and \ac{VLM}/\ac{VLA} models), and real-time capable. Success is achieved when the robot reaches the goal region or establishes a valid object-relative pose (for open-vocabulary targets), without violating safety constraints throughout the trajectory.

\subsection{Robot Dynamics and Environment Models}
We represent the  robot pose in 2D as $\mathbf{x} = [\,p_x,\, p_y,\, \psi\,]^\top \in \mathbb{R}^3$, 
where $(p_x,p_y)$ is the global position and $\psi$ the yaw. 
The control is the body-frame twist $\mathbf{u}_k = [\,v_x,\, v_y,\, \omega\,]^\top$ 
with sampling time $\Delta t$, and we adopt a constant-velocity, discrete-time dynamic model:
\begin{equation}
    \label{eq:constant_vel_model}
    \mathbf{x}_{k+1} = \mathbf{x}_{k} + \mathbf{R}(\psi_k)\,\mathbf{u}_k\,\Delta t
\end{equation}
This planar body-twist model is suitable for motion planning for both the humanoid and quadrupedal robots.

The local environment is encoded by a \ac{SDF} $\phi:\mathbb{R}^2 \to \mathbb{R}$. For any point $\mathbf{p}\in\mathbb{R}^2$, $\phi(\mathbf{p})$ returns the minimum Euclidean distance to the nearest obstacle surface; $\phi(\mathbf{p})>0$ indicates free space, $\phi(\mathbf{p})=0$ lies on an obstacle boundary, and $\phi(\mathbf{p})<0$ is inside an obstacle.

\subsection{The Proposed Caging  Framework}

Our framework employs a hierarchical control architecture that translates high-level semantic instructions into safety-assured motion (see Fig.~\ref{fig:system_pipeline}). The upper layer converts visual and language inputs into spatial navigation goals. The lower layer converts these goals into feasible control commands while enforcing safety constraints. This separation allows the semantic planner and the robot platform to be changed independently, while preserving a common safety-assured execution layer.

\subsubsection{Semantic Grounding Layer}
This layer converts visual observations and natural-language instructions into a sequence of high-level navigation actions. A \ac{VLM} processes the user's command together with the current scene and predicts discrete actions, such as turning left or right in $15^\circ$ increments up to $45^\circ$, or moving forward in $25$\,cm increments up to $75$\,cm. The \textit{Kinematic Goal Projector} maps each action to a planar goal pose $\mathbf{x}^{\mathrm{goal}}$ in the local map and verifies that the resulting waypoint is reachable and collision-free. After the mission-level action sequence is completed, an \textit{open-vocabulary object-centric grounding module} may be activated to define the robot's final approach relative to a specified object. The semantic grounding components are detailed in Section~\ref{sec:vla}.

\subsubsection{Physical Grounding Layer}
This layer converts the goal poses $\mathbf{x}^{\mathrm{goal}}$ provided by the semantic grounding layer into executable control commands while enforcing the prescribed safety constraints. For the humanoid platform, which is equipped with high-quality LiDAR, we use FastLIVO2~\cite{Zheng2024_TRO} to obtain robust LiDAR--inertial odometry. For the quadruped platform, an invariant EKF~\cite{Barrau2017TAC} fuses leg-kinematic and IMU measurements. The local environment is represented by an incrementally updated \ac{SDF}~\cite{Han2019_IROS}, which provides obstacle-clearance information directly to the motion planner. To enable real-time reactive planning with formal safety guarantees under the stated assumptions, we develop an efficient low-dimensional \ac{MPPI} planner augmented with \ac{dEBF} constraints as discussed in Section~\ref{sec:loco}. Additionlly an optional JPS global planner~\cite{Harabor2011JPS} can be included to reduce the susceptibility to local minima.

\section{High-Level VLA Planning}
\label{sec:vla}

The \textit{Kinematic Goal Projector} converts discrete high-level action commands generated by the VLA model (e.g., \texttt{turn left}, \texttt{turn right}, \texttt{move forward}) into, obstacle-free goal poses in the map. To achieve this, the module integrates odometry, and an \ac{SDF} representation of the environment. The \ac{SDF} grid is queried for the signed distance $\phi(\mathbf{p})$ at candidate position $\mathbf{p}$, which is  obstacle free if $\phi(\mathbf{p}) \geq d_{\mathrm{clear}}$, with $d_{\mathrm{clear}}=0.3\,$m as the minimum clearance threshold. If the primary position $\mathbf{p}_d$ is occupied, we select a valid alternative within a local radius which is after sampling candidate poses on the radius. Candidate poses are evaluated with a cost function that prioritizes proximity to the desired pose, alignment with the robot's heading, and obstacle clearance:
\begin{align}
C_{\text{goal}}(\mathbf{p}) = \underbrace{\|\mathbf{p} - \mathbf{p}_d\|}_{\text{proximity}}
+ w_{\psi} |\Delta \psi|
- w_{\phi} \phi(\mathbf{p})
\end{align}
where $\Delta \psi$ is the angular deviation from the forward direction and $w_{\psi}, w_{\phi}$ are weighting parameters. The valid candidate with the lowest cost is dispatched to the local planner. This design ensures the navigation module always receives goals in safe regions. Moreover, this also reduces the risk of local minima and provides more reliable final goal locations for local planners.

The \textit{open vocabulary object-centric grounding module} is activated after the mission-level language navigation is complete. Its purpose is to semantically ground the final stage of the task: detecting and approaching a user-specified target object. To achieve this, we integrate Grounding-DINO~\cite{Liu2023_arxiv} for open-vocabulary object detection and SAM2~\cite{Ravi2024_arxiv} for precise segmentation, producing both bounding boxes and fine-grained masks. The segmentation mask or bounding box centroid is then projected into 3D using the robot’s lidar, odometry and camera intrinsics, yielding a target position in the global frame.
To reject spurious detections, the target position $\mathbf{p}_t$ must satisfy 
$d_{\min} \leq \|\mathbf{p}_t - \mathbf{p}_r\| \leq d_{\max}$, 
with $d_{\min}=0.5\,$m and $d_{\max}=3.0\,$m.

The navigation state transitions through \texttt{SEARCHING}, \texttt{APPROACHING}, and \texttt{CLOSE}, depending on detection confidence and the measured target distance. When the distance falls below a threshold $d_{\text{stop}}$ (set to $0.50\text{ m}$ in our experiments), the robot halts in front of the target. This ensures the robot reaches semantically meaningful endpoints, rather than merely completing geometric navigation. 
As will be shown in our experiments, 
enabling this module allows the robot to consistently approach user-specified objects, thereby extending the system from pure language-conditioned motion planning to semantically grounded, object-aware interaction.

\section{Physical Grounding Layer}
\label{sec:loco}

The semantic layer specifies \emph{where} to move but does not account for robot geometry, obstacles, or closed-loop safety. The \emph{Physical Grounding Layer} in Fig.~\ref{fig:system_pipeline} therefore acts as an execution cage, converting each semantic goal $\mathbf{x}^{\mathrm{goal}}$ into a dynamically feasible and safe motion. This separation allows the VLM/VLA front end to remain independent of the mechanisms responsible for physical safety.

The planner combines: (i) low-dimensional \ac{MPPI}, with horizon controls parameterized by B\'{e}zier control points; (ii) an \ac{SDF}-based \ac{dEBF} cost that biases sampling toward collision-free trajectories; and (iii) a hard rollout filter with a one-step-safe backup policy. The cost improves sampling efficiency, while the filter and backup policy establish safety under the stated modeling, mapping, and tracking assumptions. Unlike collision penalties, which may trade safety for goal progress, the proposed method explicitly certifies each executed command. It also avoids numerical differentiation of an incrementally updated SDF, as required by many gradient-based \ac{MPC} formulations~\cite{Zeng2021ACC} and also avoids storage and querying of a separate gradient map. We instead evaluate it along sampled rollouts using only \ac{SDF} value queries, accommodating nonsmooth costs, nonlinear constraints, and online map updates.

\subsection{Low-Dimensional Barrier-Certified \ac{MPPI}}
\label{sec:foundations}
At each control cycle, \ac{MPPI} samples $N$ control sequences, rolls out their trajectories, and updates the sampling mean by exponential cost weighting~\cite{Williams2018_TRO}. For $\mathcal{T}^{(n)}=\{\mathbf{x}^{(n)}_k\}_{k=0}^{H}$,
$w^{(n)}\propto\exp(-J^{(n)}/\lambda)$, where $\lambda>0$ controls concentration around low-cost samples.

To reduce dimension and suppress non-smooth commands, each sequence is represented by a degree-$d$ B\'{e}zier curve~\cite{Kulkarni2026ICRA}:
\begin{equation}
    \label{eq:bezier_curve}
    \mathbf{u}(s)=\sum_{j=0}^{d}\binom{d}{j}(1-s)^{d-j}s^jP_j,
    \qquad s\in[0,1],
\end{equation}
where $\{P_j\}_{j=0}^{d}$ are control points. The optimization thus uses $d+1$ control points instead of one control vector per horizon step and produces smooth commands by construction. After certification, only the first command is executed before replanning.
The rollout objective is
\begin{equation}
    \label{eq:Jn}
    J^{(n)}=C_\mathrm{goal}^{(n)}+C_\mathrm{ctl}^{(n)}+C_\mathrm{safety}^{(n)},
\end{equation}
where the first two terms rank goal progress and command quality, while $C_\mathrm{safety}^{(n)}$ shapes sampling but does not provide the safety guarantee.

\subsection{Goal-Tracking and Control Costs}

Let $\mathbf{p}_k$ and $\mathbf{p}^{\mathrm{goal}}$ be the current and desired planar positions, with
$\ell_\mathrm{pos}(\mathbf{x}_k)=\|\mathbf{p}_k-\mathbf{p}^{\mathrm{goal}}\|_2$.
Yaw tracks motion direction far from the goal and terminal yaw $\psi^{\mathrm{goal}}$ near it
\begin{align}
    \alpha_\psi
    =\sigma\!\left(
    \frac{\|\mathbf{p}_k-\mathbf{p}^{\mathrm{goal}}\|_2-d_\mathrm{switch}}
    {s_\mathrm{switch}}
    \right),
\end{align}
where $d_\mathrm{switch}$ and $s_\mathrm{switch}$ set the transition location and sharpness, and
\begin{align}
    \ell_\mathrm{yaw}(\mathbf{x}_k)
    ={}&\alpha_\psi|\psi_k-\psi_k^{\mathrm{head}}|
    +(1-\alpha_\psi)|\psi_k-\psi^{\mathrm{goal}}|,
\end{align}
with $\psi_k^{\mathrm{head}}$ induced by predicted motion. Thus,
\begin{equation}
    C_\mathrm{goal}=\sum_{k=0}^{H-1}
    \left(w_\mathrm{pos}\ell_{\mathrm{pos},k}
    +w_\mathrm{yaw}\ell_{\mathrm{yaw},k}\right).
\end{equation}
Command magnitude and componentwise limit violations are penalized by
\begin{align}
    C_\mathrm{ctl}=\sum_{k=0}^{H-1}\left(
    w_\mathrm{ctl}\|\mathbf{u}_k\|_2
    +\sum_{q=1}^{3}w_\mathrm{lim}
    [|u_{k,q}|-u_{q,\max}]_+\right),
\end{align}
where $[z]_+=\max(z,0)$. Together with the B\'{e}zier representation, this discourages aggressive commands and encourages compliance with platform velocity limits.

\subsection{SDF-Based dEBF Safety}
\label{subsec:multi_point_barrier}
For barrier function $h:\mathbb{R}^n\rightarrow\mathbb{R}$, define the safe set
\begin{equation}
    \label{eq:safe_set}
    \mathcal{S}\triangleq\{\mathbf{x}\in\mathbb{R}^n\mid h(\mathbf{x})\geq0\}.
\end{equation}
The discrete exponential barrier condition~\cite{Agrawal2017RSS}
\begin{equation}
    \label{eq:dbf_step}
    h(\mathbf{x}_{k+1})\geq\gamma h(\mathbf{x}_k),
\end{equation}
with $\gamma\in[0,1)$, limits the rate of approach to the boundary: as $h(\mathbf{x}_k)\rightarrow0$, the allowable next-step decrease vanishes. Repeated satisfaction preserves nonnegative barriers and makes $\gamma$ an interpretable conservatism parameter, with larger values retaining more clearance.

\begin{figure}
    \centering
    \includegraphics[width=0.8\linewidth]{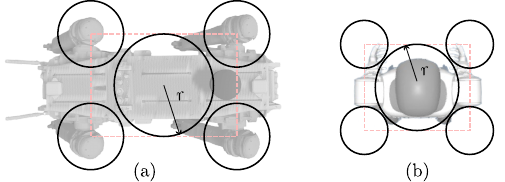}
    \caption{Multi-point safety models for (a) Ghost Robotics Vision~60 and (b) Unitree G1. Each point has a radius representing the local footprint.}
    \label{fig:multi-point}
\end{figure}

We approximate the robot by $M$ body-fixed spheres of radii $r_i\geq0$ (Fig.~\ref{fig:multi-point}). The map $g_i(\mathbf{x}_k)$ gives sphere center $\mathbf{p}_{i,k}\in\mathbb{R}^2$, and the SDF $\phi$ gives clearance
$c_i(\mathbf{x}_k)\triangleq\phi(\mathbf{p}_{i,k})-r_i$.
Setting $h_i(\mathbf{x}_k)\triangleq c_i(\mathbf{x}_k)$ yields
\begin{equation*}
    \mathcal{S}\triangleq\bigcap_{i=1}^{M}\mathcal{S}_i,
    \qquad
    \mathcal{S}_i\triangleq\{\mathbf{x}\mid h_i(\mathbf{x})\geq0\}.
\end{equation*}
The robot is collision-free iff all barriers are nonnegative and both the platforms are captured with few SDF queries.

During optimization, predicted dEBF violations are penalized to favor candidates likely to pass hard certification:
\begin{align}
    \label{eq:relu_dbf}
    C_\mathrm{safety}(\mathcal{T}^{(n)})
    =\sum_{k=0}^{H-1}\sum_{i=1}^{M}w_i
    \Big[\gamma h_i(\mathbf{x}^{(n)}_k)
    -h_i(\mathbf{x}^{(n)}_{k+1})
    +\varepsilon_\mathrm{bar}\Big]_+,
\end{align}
where $w_i$ weights geometric points and $\varepsilon_\mathrm{bar}>0$ adds a planning margin. This cost guides sampling; safety is enforced by explicit constraint checking.

\begin{algorithm}[t]
\caption{Barrier-Certified Low-Dimensional MPPI}
\label{alg:safe_mppi_update}
\begin{algorithmic}[1]
\State \textbf{Input:} $\mathbf{x}_0$, $\lambda$, $N_\mathrm{samples}$, $N_\mathrm{iters}$, $\kappa(\cdot)$
\State Initialize B\'{e}zier control points $P$
\For{$m=1$ to $N_\mathrm{iters}$}
    \State Sample $P^{(j)}\sim\mathcal{N}(P,\Sigma)$, $j=1,\ldots,N_\mathrm{samples}$
    \For{$j=1$ to $N_\mathrm{samples}$}
        \State Construct $\mathbf{u}^{(j)}$ using~\eqref{eq:bezier_curve}
        \State Roll out~\eqref{eq:constant_vel_model} from $\mathbf{x}_0$
        \State Evaluate $J^{(j)}$ and $\tilde{w}^{(j)}\gets\exp(-J^{(j)}/\lambda)$
    \EndFor
    \State $w^{(j)}\gets\tilde{w}^{(j)}/\sum_{\ell=1}^{N_\mathrm{samples}}\tilde{w}^{(\ell)}$
    \State $P\gets\sum_{j=1}^{N_\mathrm{samples}}w^{(j)}P^{(j)}$
\EndFor
\State $\mathcal{I}\gets\operatorname{argsort}_{j}J^{(j)}$
\For{$j\in\mathcal{I}$}
    \If{\eqref{eq:dbf_step} holds $\forall i,k$ along rollout $j$}
        \State \Return first command $\mathbf{u}^{(j)}_0$
    \EndIf
\EndFor
\State \Return $\kappa(\mathbf{x}_0)$
\end{algorithmic}
\end{algorithm}

The controller first uses~\eqref{eq:relu_dbf} to bias sampling toward safe regions, then executes the lowest-cost rollout satisfying~\eqref{eq:dbf_step} for every geometric point and predicted transition. If none is admissible, it applies a one-step-safe backup $\kappa(\mathbf{x}_k)$ satisfying
\begin{equation}
    \label{eq:backup}
    h_i\!\left(
    \mathbf{x}_k+\mathbf{R}(\psi_k)\kappa(\mathbf{x}_k)\Delta t
    \right)
    \geq\gamma h_i(\mathbf{x}_k),
    \quad\forall i.
\end{equation}

\begin{thm}[Safety of dEBF-Certified MPPI]
\label{thm:dbf_mppi_safe}
Assume: (A1) the state follows~\eqref{eq:constant_vel_model}; (A2) valid robot geometry and the SDF define $\mathcal{S}=\bigcap_{i=1}^{M}\{\mathbf{x}\mid h_i(\mathbf{x})\geq0\}$; (A3) for every $\mathbf{x}_k\in\mathcal{S}$, a one-step backup $\kappa(\mathbf{x}_k)$ exists and satisfies~\eqref{eq:backup}; (A4) the controller executes the first sampled command only if its transition satisfies~\eqref{eq:dbf_step} for every barrier, otherwise apply the backup; and (A5) $\mathbf{x}_0\in\mathcal{S}$. Then, for $k\in\mathbb{N}_0$ and  $\gamma\in[0,1)$,
\begin{equation}
    \label{eq:theorem_safety_bound}
    h_i(\mathbf{x}_k)\geq\gamma^k h_i(\mathbf{x}_0)\geq0,
    \quad \forall i\in\{1,\ldots,M\},
\end{equation}
so 
$\mathcal{S}$ is forward invariant.
\end{thm}

The proof is in Appendix~\ref{app:proof}. For~\eqref{eq:constant_vel_model}, zero body velocity is a valid backup from any safe state, making the planning-level certification recursively feasible.
Unlike collision penalty-only \ac{MPPI}, the dEBF couples consecutive clearances and regulates approach rate rather than only penalizing penetration and thus resulting a longer \emph{"lookahead"} with a shorter horizon \cite{Zeng2021ACC}. Unlike gradient-based SDF-constrained \ac{MPC}, it requires no map derivatives; unlike CBF-in-MPPI methods with a separate refinement stage~\cite{Yin2023_arxiv}, it certifies sampled rollouts directly. The margin $\varepsilon_\mathrm{bar}$ improves candidate robustness, while the hard filter and backup enforce the constraint. The resulting front-end-agnostic safety cage reacts to newly observed geometry and its evaluations are presented Sec.~\ref{sec:exp}.

\section{Experimental Validation}
\label{sec:exp}

We structure the evaluation around three questions. \textbf{(Q1)} How does the
proposed \ac{dEBF} affect obstacle clearance and object-approach behavior relative to
standard \ac{MPPI}? \textbf{(Q2)} Does dEBF reduce observed collision incidence
both without a global path planner and when JPS-based global
replanning~\cite{Harabor2011JPS} is available? \textbf{(Q3)} Does the complete
physical-grounding cage remain effective across environment scales, semantic
front ends, and legged platforms? We first characterize dEBF in simulation,
then isolate its effect on a Unitree G1 humanoid, and finally evaluate
end-to-end navigation on the G1 and a Ghost Robotics Vision~60 quadruped.

\subsection{Experimental Setup and Metrics}

\noindent\textbf{Simulation.}
We use Gazebo to simulate a Vision~60, LiDAR, and ground-truth odometry.
The mapping and perception pipeline mirrors the hardware implementation. The goal is held fixed so that the experiments isolate the local planner
under different dEBF parameters and obstacle configurations.

\noindent\textbf{Hardware.}
The G1 and Vision~60 carry an RGB camera, IMU, and LiDAR for online state
estimation, mapping, and geometric grounding. Platform-specific locomotion
controllers track the commanded body-frame velocity, while an NVIDIA A4500
executes the mapping and motion-planning stack. The semantic front ends are
executed locally or accessed through their respective interfaces.

\noindent\textbf{Planner parameters.}
For all experiments, the dEBF decay parameter is set to $\gamma=0.95$.
At each control update, MPPI uses 200 samples and 10 optimization iterations.
Each rollout is discretized with a $0.05\,\mathrm{s}$ step over a horizon of
50 time steps, corresponding to a $2.5\,\mathrm{s}$ prediction horizon.

The G1 is the primary controlled-evaluation platform. We conduct physical grounding layer
ablations in a $4\,\mathrm{m}\times4\,\mathrm{m}$ arena, an end-to-end study in
a $6\,\mathrm{m}\times6\,\mathrm{m}$ static-clutter environment, and a
system-level caging study in a $12\,\mathrm{m}\times12\,\mathrm{m}$
environment. The controlled studies compare MPPI with and without dEBF across
ChatGPT, Claude, Gemini, and NaVILA. The moderate-scale study uses 25
instructions, each repeated three times under both planner configurations,
yielding 75 trials per configuration. The large-scale study evaluates 15
instructions once with and once without the complete the physical-grounding cage for each of the four
semantic front ends, yielding 120 executions. The Vision~60 experiments assess
cross-platform deployment in real indoor and outdoor environments.

\noindent\textbf{Metrics.}
Success Rate (SR) is the fraction of trials that complete the instructed
navigation task. Collision Rate (CR) is the fraction in which the robot
contacts an obstacle or violates the predefined collision boundary.
End-to-End Success Rate (E2E SR) requires both task completion and
collision-free execution. For object-centric navigation, Object SR denotes
successful localization and approach, and Navigation Error (NE) is the final
distance to the desired object-relative terminal position, following standard
VLN practice~\cite{Cheng2025}. We report only metrics annotated for each study
and do not infer unrecorded task success from collision outcomes.

\subsection{Simulation Characterization of dEBF}

\begin{figure*}[t]
    \centering
    \includegraphics[width=0.9\textwidth,trim=10 10 0 0,clip]
    {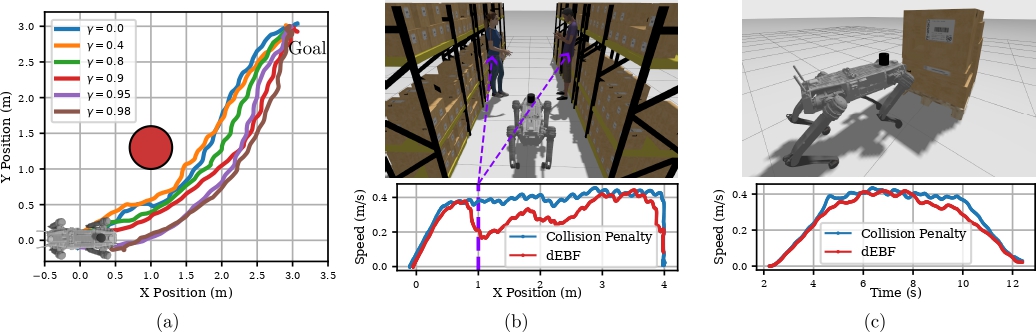}
    \caption{Qualitative effects of dEBF on clearance and approach-rate
    regulation. (a) Effect of $\gamma$ on obstacle clearance. (b) Speed
    modulation in a narrow passage for collision-penalty MPPI and MPPI+dEBF.
    (c) Smooth final approach in a object approach.}
    \label{fig:evaluating_planner}
    \vspace{-10pt}
\end{figure*}

Figure~\ref{fig:evaluating_planner} illustrates how dEBF shapes local motion.
Increasing the decay parameter $\gamma$ requires a predicted trajectory to
retain a larger fraction of its current barrier value~\eqref{eq:dbf_step} and therefore produces
more conservative motion around an obstacle
(Fig.~\ref{fig:evaluating_planner}(a)). In a narrow passage, the standard
collision-penalty MPPI maintains a relatively high speed until it approaches the obstacle
boundary, whereas MPPI+dEBF slows as predicted clearance decreases
(Fig.~\ref{fig:evaluating_planner}(b)). The object-approach example similarly
shows a smooth reduction in speed during the final approach
(Fig.~\ref{fig:evaluating_planner}(c)).

These behaviors arise from the temporal coupling in the discrete barrier
condition. A conventional collision penalty evaluates individual predicted
states and may trade obstacle proximity against goal progress. In contrast,
dEBF constrains the permitted decrease in clearance between consecutive
predicted states, regulating both intersection with the safe-set boundary and
the rate at which a rollout approaches it. Making dEBF+MPPI highly suitable for safe and smooth navigation through cluttered and narrow spaces.

\subsection{Controlled dEBF Ablations on G1}
\label{sec:g1_controlled}

We conduct two complementary studies in the
$4\,\mathrm{m}\times4\,\mathrm{m}$ G1 arena. In the first, JPS is disabled and
a fixed obstacle interferes with the nominal direction to a target object.
Thus, avoidance is determined by local MPPI, with or without dEBF. In the
second, JPS is enabled in both configurations and a previously unobserved
obstacle is suddenly introduced after the robot begins moving. The obstacle is
incorporated into the online map and triggers global replanning. This paired
design evaluates whether dEBF adds short-horizon protection during perception,
map updating, and replanning rather than comparing local and global planning.
Initial pose, obstacle and target placement, velocity limits, planner
parameters, and timeout are held fixed within each comparison.

Tables~\ref{tab:g1_no_jps_hybrid} and~\ref{tab:g1_jps_hybrid} show a
consistent collision reduction in both settings. Without JPS, the aggregate CR
decreases from 80\% to 0\%; with JPS shared by both configurations, it
decreases from 100\% to 0\%. No collisions were observed in any of the 40
dEBF-enabled controlled trials. The aggregate Object SR changes from 35\% to
40\% in the static study and from 55\% to 70\% in the
sudden-obstacle-introduction study. The corresponding aggregate E2E SR
increases from 15\% to 40\% and from 0\% to 70\%, respectively.
JPS and dEBF serve complementary roles. JPS supplies global route guidance
around mapped obstacles, reducing local MPPI's susceptibility to local minima;
dEBF constrains short-horizon rollouts during map-update and replanning
transitions.

The consistent collision reduction shows that dEBF improves physical execution
across all four tested semantic models. The proposed Legged-VLN architecture
physically grounds the model's navigation proposals but does not modify its
semantic reasoning. Task completion therefore remains dependent on the
underlying model; for example, the Claude trials in the no-JPS setting are
collision-free with dEBF yet do not complete the object-approach task. Thus,
these results demonstrate improved collision avoidance without implying
improved semantic reasoning or target selection.

\subsection{End-to-End Scaling and Cross-Platform Transfer}
\label{sec:end_to_end_hybrid}

We next evaluate longer language-conditioned missions. On G1, the
$6\,\mathrm{m}\times6\,\mathrm{m}$ environment contains multiple static
obstacles and tasks ranging from one- or two-action instructions to longer
sequences involving multiple landmarks and turns. Each of 25 instructions is
executed three times under standard MPPI and MPPI+dEBF, yielding 75 trials per
configuration. On Vision~60, we use the same planning-level interface in real
indoor and outdoor environments. State estimation, sensor coverage, footprint
geometry, and low-level locomotion remain platform specific; in particular a platform-specific
multi-point footprint is used for SDF queries as in Fig~\ref{fig:multi-point}.

\begin{table*}[t]
    \centering
    \begin{minipage}[t]{0.32\textwidth}
        \centering
        \captionof{table}{Static-obstacle G1 evaluation with JPS disabled.
        Results compare dEBF Off/On using five trials per front end and 20
        aggregate trials per planner configuration.}
        \label{tab:g1_no_jps_hybrid}
        \setlength{\tabcolsep}{2.4pt}
        \resizebox{\linewidth}{!}{
        \begin{tabular}{lcccc}
            \toprule
            Front End & $N$ &
            \shortstack{CR\\Off/On} &
            \shortstack{Obj. SR\\Off/On} &
            \shortstack{E2E SR\\Off/On} \\
            \midrule
            Gemini  & 5 & 100/\textbf{0}\% & 40/\textbf{80}\%
                    & 0/\textbf{80}\% \\
            ChatGPT & 5 & 80/\textbf{0}\% & 0/\textbf{40}\%
                    & 0/\textbf{40}\% \\
            Claude  & 5 & 60/\textbf{0}\% & 80/0\%
                    & 40/0\% \\
            NaVILA  & 5 & 80/\textbf{0}\% & 20/\textbf{40}\%
                    & 20/\textbf{40}\% \\
            \midrule
            Aggregate & 20 & 80/\textbf{0}\% & 35/40\%
                    & 15/\textbf{40}\% \\
            \bottomrule
        \end{tabular}}
    \end{minipage}
    \hfill
    \begin{minipage}[t]{0.32\textwidth}
        \centering
        \captionof{table}{G1 with JPS enabled and a suddenly
        introduced obstacle.
        Results compare dEBF Off/On using 5 trials per front end and 20
        aggregate trials per planner configuration.}
        \label{tab:g1_jps_hybrid}
        \setlength{\tabcolsep}{2.4pt}
        \resizebox{\linewidth}{!}{
        \begin{tabular}{lcccc}
            \toprule
            Front End & $N$ &
            \shortstack{CR\\Off/On} &
            \shortstack{Obj. SR\\Off/On} &
            \shortstack{E2E SR\\Off/On} \\
            \midrule
            Gemini  & 5 & 100/\textbf{0}\% & 100/100\%
                    & 0/\textbf{100}\% \\
            ChatGPT & 5 & 100/\textbf{0}\% & 60/\textbf{100}\%
                    & 0/\textbf{100}\% \\
            Claude  & 5 & 100/\textbf{0}\% & 40/\textbf{60}\%
                    & 0/\textbf{60}\% \\
            NaVILA  & 5 & 100/\textbf{0}\% & 20/20\%
                    & 0/\textbf{20}\% \\
            \midrule
            Aggregate & 20 & 100/\textbf{0}\% & 55/\textbf{70}\%
                    & 0/\textbf{70}\% \\
            \bottomrule
        \end{tabular}}
    \end{minipage}
    \hfill
    \begin{minipage}[t]{0.32\textwidth}
        \centering
        \captionof{table}{End-to-end G1 and Vision~60 evaluation with MPPI and
        MPPI+dEBF. Results use 75 G1 trials per planner and the available
        Vision~60 category-level rates.}
        \label{tab:main_realworld_hybrid}
        \setlength{\tabcolsep}{7pt}
        \resizebox{\linewidth}{!}{
        \begin{tabular}{lcccc}
            \toprule
            & \multicolumn{2}{c}{Simple} &
            \multicolumn{2}{c}{Complex} \\
            \cmidrule(lr){2-3}\cmidrule(lr){4-5}
            Method & SR $\uparrow$ & CR $\downarrow$
            & SR $\uparrow$ & CR $\downarrow$ \\
            \midrule
            \multicolumn{5}{c}{\textit{Unitree G1}} \\
            MPPI & 0.75 & 0.33 & 0.75 & 0.50 \\
            MPPI+dEBF
            & \textbf{0.75} & \textbf{0.00}
            & \textbf{0.75} & \textbf{0.00} \\
            \midrule
            \multicolumn{5}{c}{\textit{Ghost Robotics Vision~60}} \\
            MPPI & 0.47 & 0.67 & 0.46 & 0.91 \\
            MPPI+dEBF
            & \textbf{0.73} & \textbf{0.20}
            & \textbf{0.54} & \textbf{0.08} \\
            \bottomrule
        \end{tabular}}
    \end{minipage}
\end{table*}

As shown in Table~\ref{tab:main_realworld_hybrid}, dEBF reduces observed CR on
both platforms. On G1, CR decreases from 0.33 to 0.00 for simple tasks and
from 0.50 to 0.00 for complex tasks, while SR remains 0.75. The unchanged SR
indicates that the reduction was not obtained by suppressing a larger fraction
of task executions. On Vision~60, dEBF reduces CR from 0.67 to 0.20 for simple
tasks and from 0.91 to 0.08 for complex tasks while increasing the recorded SR.

The remaining Vision~60 collisions are primarily associated with incomplete
environmental representation, including thin wires, table edges, narrow poles,
and geometry in sensor blind spots due to an older LiDAR VLP16. Since the barrier is evaluated against the
current SDF, absent geometry cannot be certified. Overall, introduction of the proposed caging architecture enhance the safety of the commands proposed by VLM/VLA models.

\begin{figure*}[t]
    \centering
    \begin{subfigure}{0.85\linewidth}
        \centering
        \includegraphics[width=\linewidth,trim=0 100 0 5,clip]
        {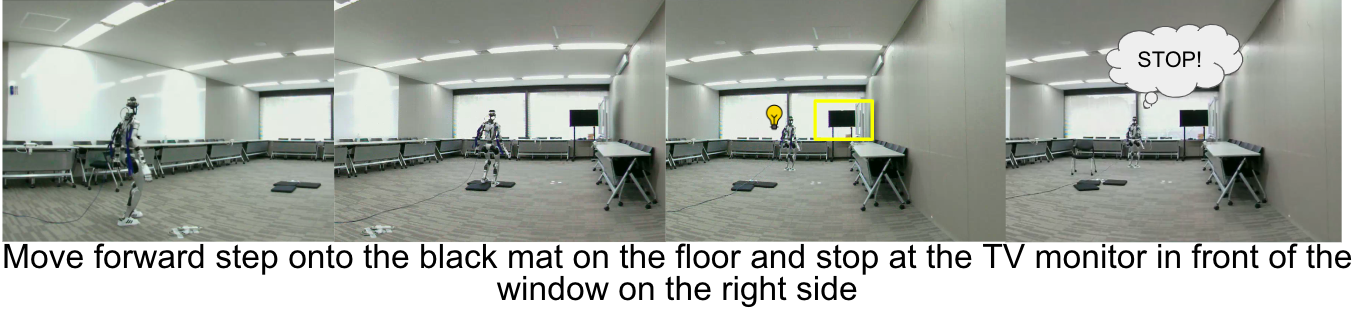}
        \caption{Unitree G1: ``Move forward, step onto the black mat, and stop
        at the TV monitor in front of the window on the right.''}
    \end{subfigure}\\[0.35em]
    \begin{subfigure}{0.85\linewidth}
        \centering
        \includegraphics[width=\linewidth,trim=0 50 0 5,clip]
        {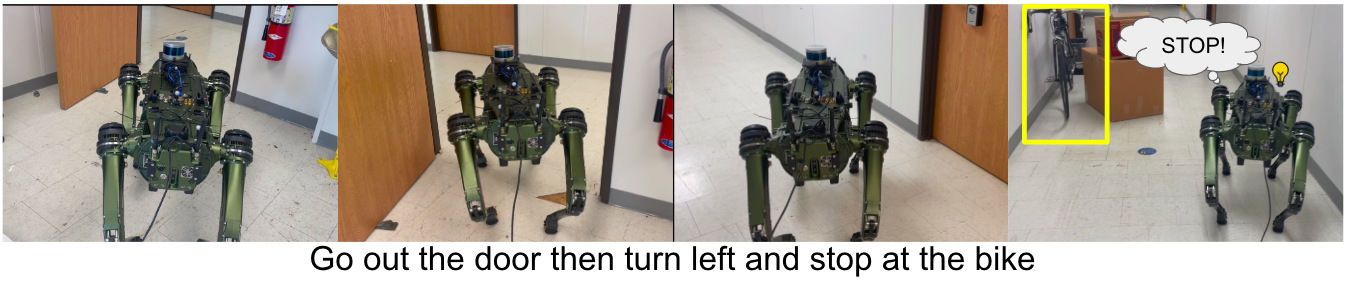}
        \caption{Vision~60: ``Go out the door, turn left, and stop at the
        bike.''}
    \end{subfigure}
    \caption{Representative language-conditioned executions on the humanoid
    and quadruped platforms.}
    \label{fig:visual_results_main}
    \vspace{-10pt}
\end{figure*}

\subsection{System-Level Caging Across Semantic Front Ends}
\label{sec:g1_large_scale}

The final G1 safety study uses a $12\,\mathrm{m}\times12\,\mathrm{m}$ indoor
environment and 15 instructions ranging from single-stage commands to
multi-stage tasks with multiple landmarks and spatial constraints. For each
front end, the same instructions are executed with the complete
physical-grounding cage disabled and enabled. Caging enables MPPI, JPS, and
dEBF jointly. In uncaged they are turned off to clearly demonstrate the benefits of the proposed caging architecture.
\begin{table}[t]
    \centering
    \caption{System-level G1 caging ablation. Each entry reports
    collisions/trials (CR).}
    \label{tab:g1_large_scale_caging}
    \setlength{\tabcolsep}{5pt}
    \begin{tabular}{lcc}
        \toprule
        Front End & Caging Off & Caging On \\
        \midrule
        ChatGPT 5.6 Luna & 15/15 (100\%) & \textbf{0/15 (0\%)} \\
        Gemini 3.5 Flash & 15/15 (100\%) & \textbf{0/15 (0\%)} \\
        Claude Fable 5   & 15/15 (100\%) & \textbf{0/15 (0\%)} \\
        NaVILA           & 15/15 (100\%) & \textbf{0/15 (0\%)} \\
        \midrule
        Aggregate        & 60/60 (100\%) & \textbf{0/60 (0\%)} \\
        \bottomrule
    \end{tabular}
\end{table}
Table~\ref{tab:g1_large_scale_caging} summarizes 120 executions. All of the 60 trails without the proposed caging architecture results in collisions for these longer tasks, whereas no collision is observed with the caging architecture. The same trend occurs within every semantic frontend VLM/VLA model, showing that caging is a necessity for all the models tested with longer tasks.

\subsection{Object-Centric Grounding and Evaluation Scope}
\label{sec:vision60_object_grounding}
\begin{table}[t]
    \centering
    \caption{Object-centric terminal grounding on Vision~60.}
    \label{tab:vision60_object_grounding}
    \setlength{\tabcolsep}{6pt}
    \begin{tabular}{lcc}
        \toprule
        Grounding & Evaluated Endpoints & Mean NE (m) $\downarrow$ \\
        \midrule
        Disabled & 5 & 1.50 \\
        Enabled  & 3 & \textbf{0.29} \\
        \bottomrule
    \end{tabular}
\end{table}
Collision-aware execution does not ensure that a mission terminates at the
intended semantic object. We therefore evaluate an auxiliary terminal-grounding
stage on Vision~60. After mission-level navigation, Grounding-DINO and SAM2
detect and segment the language-specified target. Its estimated position is
projected into the global frame, and the robot performs a final object-relative
approach. Table~\ref{tab:vision60_object_grounding} reports a decrease in mean endpoint
error from 1.50\,m to 0.29\,m with object-centric grounding. This layer is beneficial for missions that require approaching an object.

\subsection{Discussion and Limitations}
The controlled ablations in Sec.~\ref{sec:g1_controlled} and end-to-end
evaluation in Sec.~\ref{sec:end_to_end_hybrid} show that uncaged collision risk
increases with task length and complexity. This trend culminates in the
long-horizon G1 ablation (Sec.~\ref{sec:g1_large_scale}), where all 60 uncaged
trials collide, whereas caging reduces the aggregate CR to 0\% across four
semantic models. JPS provides global route guidance around mapped obstacles,
reducing the local planner's susceptibility to local minima, but does not
certify local execution; with JPS enabled, adding dEBF reduces CR from 100\% to
0\% (Table~\ref{tab:g1_jps_hybrid}). Similar improvements on Vision~60 support
portability across the tested task scales, environments, and platforms.

These results remain subject to the assumptions of
Theorem~\ref{thm:dbf_mppi_safe}.
Certification uses the current SDF and cannot account for unobserved geometry.
The planar body-twist and multi-point models also exclude state-estimation error,
latency, and tracking error. Object-centric grounding
(Sec.~\ref{sec:vision60_object_grounding}) improves terminal alignment, but the
cage cannot correct erroneous or unreachable semantic goals.

\section{Conclusions and Future Work}

We presented a modular caging architecture for legged \ac{VLN} that separates
semantic reasoning from physically grounded execution. High-level
\ac{VLM}/\ac{VLA} outputs are treated as navigation proposals and converted
into geometric goals, while a low-dimensional B\'ezier-parameterized
\ac{MPPI} planner uses the online \ac{SDF} and \ac{dEBF} constraints to
generate executable motion. This separation permits different semantic front
ends and legged platforms to share the same planning-level safety interface.
We prove collision avoidance for dEBF-constrained MPPI under the stated assumptions. We extensively evaluated the proposed Legged-\ac{VLN} architecture on a
Unitree G1 humanoid and a Ghost Robotics Vision~60 quadruped. The real-world
experiments demonstrate modularity across platforms and substantially reduced
observed collision incidence, while open-vocabulary object grounding improves
semantic terminal alignment under the tested conditions. Future work will incorporate uncertainty in state estimation and mapping, as
well as moving obstacles, into the safety analysis to narrow the gap between
the formal model and hardware execution. We will also extend the capabilities
of \ac{VLA} models beyond navigation to richer humanoid skills, including
grasping, object handover, and fine-grained target interaction. Finally,
integrating multimodal memory will support longer-horizon reasoning toward
general-purpose robotic assistants.

\appendices

\section{Proof of Safe Motion Plan}
\label{app:proof}

\begin{proof}[Theorem~\ref{thm:dbf_mppi_safe}]
By induction on the control step \(k\).

\emph{Base case:} By Assumption~(A5), \(\mathbf{x}_0\in\mathcal{S}\), so \(h_i(\mathbf{x}_0)\geq0\) for every \(i\). Since \(\gamma^0=1\), we have 
$
    h_i(\mathbf{x}_0)
    =\gamma^0 h_i(\mathbf{x}_0)
    \geq0.
$
Thus,~\eqref{eq:theorem_safety_bound} holds at \(k=0\).

\emph{Inductive step:} Suppose that, at some step \(k\),
\begin{equation*}
    h_i(\mathbf{x}_k)
    \geq\gamma^k h_i(\mathbf{x}_0)
    \geq0,
    \qquad \forall i.
\end{equation*}
By Assumption~(A4), the controller applies one of two inputs. If a certified MPPI candidate exists, it executes a first command whose resulting transition satisfies~\eqref{eq:dbf_step} for every \(i\). Otherwise, it applies the backup input \(\mathbf{u}_k=\kappa(\mathbf{x}_k)\), which satisfies the same one-step condition by Assumption~(A3) and~\eqref{eq:backup}. Hence, in either case,
\begin{equation*}
    h_i(\mathbf{x}_{k+1})
    \geq\gamma h_i(\mathbf{x}_k),
    \qquad \forall i.
\end{equation*}
Applying the inductive hypothesis for all $i$ gives
\begin{equation*}
    h_i(\mathbf{x}_{k+1})
    \geq\gamma h_i(\mathbf{x}_k) 
    \geq\gamma\bigl(\gamma^k h_i(\mathbf{x}_0)\bigr)
    =\gamma^{k+1}h_i(\mathbf{x}_0) \geq0,
\end{equation*}
where the final inequality follows from \(\gamma\in[0,1)\) and \(h_i(\mathbf{x}_0)\geq0\). Therefore,~\eqref{eq:theorem_safety_bound} holds at step \(k+1\). By induction, every barrier remains nonnegative for all \(k\in\mathbb{N}_0\). Thus \(\mathbf{x}_k\in\mathcal{S}\) for all time, and the safe set is forward invariant.
\end{proof}

{
\bibliographystyle{packages/IEEEtran}
\bibliography{libraries/rpng,libraries/vln,libraries/legged, libraries/related, libraries/extra}

@INPROCEEDINGS{Macenski2020IROS,
  author={Macenski, Steve and Martín, Francisco and White, Ruffin and Clavero, Jonatan Ginés},
  booktitle={2020 IEEE/RSJ International Conference on Intelligent Robots and Systems (IROS)}, 
  title={The Marathon 2: A Navigation System}, 
  year={2020},
  volume={},
  number={},
  pages={2718-2725},
  doi={10.1109/IROS45743.2020.9341207}}

@INPROCEEDINGS{Kulkarni2026ICRA,
  author={Kulkarni, Abhijeet M. and Poulakakis, Ioannis and Huang, Guoquan},
  booktitle={2026 IEEE International Conference on Robotics and Automation (ICRA)}, 
  title={Learning Neural Observer-Predictor Models for Limb-level Sampling-based Locomotion Planning}, 
  year={2026},
  volume={},
  number={},
  pages={21515-21522},
  doi={10.1109/ICRA57385.2026.11697171}}

@ARTICLE{Narkhede2022RAL,
  author={Narkhede, Kunal S. and Kulkarni, Abhijeet M. and Thanki, Dhruv A. and Poulakakis, Ioannis},
  journal={IEEE Robotics and Automation Letters}, 
  title={A Sequential MPC Approach to Reactive Planning for Bipedal Robots Using Safe Corridors in Highly Cluttered Environments}, 
  year={2022},
  volume={7},
  number={4},
  pages={11831--11838},
  doi={10.1109/LRA.2022.3204367}}

@article{Chen2025TRO,
  author  = {Chen, Timothy and Shorinwa, Ola and Bruno, Joseph and Swann, Aiden and Yu, Javier and Zeng, Weijia and Nagami, Keiko and Dames, Philip M. and Schwager, Mac},
  title   = {Splat-Nav: Safe Real-Time Robot Navigation in Gaussian Splatting Maps},
  journal = {IEEE Transactions on Robotics},
  year    = {2025},
  volume  = {41},
  pages   = {2765--2784},
  month   = mar,
  doi     = {10.1109/TRO.2025.3552348}
}

@inproceedings{Agrawal2017RSS,
  title={Discrete Control Barrier Functions for Safety-Critical Control of Discrete Systems with Application to Bipedal Robot Navigation},
  author={Ayush Agrawal and K. Sreenath},
  booktitle={Proc. Robot. Sci. Syst.},
  year={2017}
}

@INPROCEEDINGS{Zeng2021ACC,
  author={Zeng, Jun and Zhang, Bike and Sreenath, Koushil},
  booktitle={Proc. Amer. Control Conf.}, 
  title={Safety-Critical Model Predictive Control with Discrete-Time Control Barrier Function}, 
  year={2021},
  volume={},
  number={},
  pages={3882-3889},
  doi={10.23919/ACC50511.2021.9483029}}

@misc{unitree2025a2_video,
  author = {{Unitree Robotics}},
  title = {{Unitree A2, a Quadruped Robotic Dog with Industrial Applications [Video]}},
  howpublished = {YouTube video},
  year = {2025},
  note = {Launched in Hangzhou on August 5, 2025}
}

@article{Barrau2017TAC,
  title={The invariant extended Kalman filter as a stable observer},
  author={Barrau, Axel and Bonnabel, Silv{\`e}re},
  journal={IEEE Transactions on Automatic Control},
  volume={62},
  number={4},
  pages={1797--1812},
  year={2017},
  publisher={IEEE}
}

@misc{Ravi2024_arxiv,
      title={SAM 2: Segment Anything in Images and Videos}, 
      author={Nikhila Ravi and Valentin Gabeur and Yuan-Ting Hu and Ronghang Hu and Chaitanya Ryali and Tengyu Ma and Haitham Khedr and Roman Rädle and Chloe Rolland and Laura Gustafson and Eric Mintun and Junting Pan and Kalyan Vasudev Alwala and Nicolas Carion and Chao-Yuan Wu and Ross Girshick and Piotr Dollár and Christoph Feichtenhofer},
      year={2024},
      eprint={2408.00714},
      archivePrefix={arXiv},
      primaryClass={cs.CV}
}

@article{Liu2023_arxiv,
  title={Grounding dino: Marrying dino with grounded pre-training for open-set object detection},
  author={Liu, Shilong and Zeng, Zhaoyang and Ren, Tianhe and Li, Feng and Zhang, Hao and Yang, Jie and Li, Chunyuan and Yang, Jianwei and Su, Hang and Zhu, Jun and others},
  journal={arXiv preprint arXiv:2303.05499},
  year={2023}
}

@inproceedings{Han2019_IROS,
  title={Fiesta: Fast incremental euclidean distance fields for online motion planning of aerial robots},
  author={Han, Luxin and Gao, Fei and Zhou, Boyu and Shen, Shaojie},
  booktitle={2019 IEEE/RSJ International Conference on Intelligent Robots and Systems (IROS)},
  pages={4423--4430},
  year={2019},
  organization={IEEE}
}

@article{Williams2018_TRO,
  author    = {Grady Williams and Andrew Aldrich and Evangelos A. Theodorou},
  title     = {Information-Theoretic Model Predictive Control: Theory and Applications to Autonomous Driving},
  journal   = {IEEE Transactions on Robotics},
  volume    = {34},
  number    = {6},
  pages     = {1603--1622},
  year      = {2018},
  doi       = {10.1109/TRO.2018.2858280}
}

@article{Yin2023_arxiv,
      title={Shield Model Predictive Path Integral: A Computationally Efficient Robust MPC Approach Using Control Barrier Functions}, 
      author={Ji Yin and Charles Dawson and Chuchu Fan and Panagiotis Tsiotras},
      year={2023},
      journal={arxiv:2302.11719},
      archivePrefix={arXiv},
      primaryClass={cs.RO}
}

@article{Parwana2025_arxiv,
      title={BR-MPPI: Barrier Rate guided MPPI for Enforcing Multiple Inequality Constraints with Learned Signed Distance Field}, 
      author={Hardik Parwana and Taekyung Kim and Kehan Long and Bardh Hoxha and Hideki Okamoto and Georgios Fainekos and Dimitra Panagou},
      year={2025},
      journal={arxiv:2506.07325},
      archivePrefix={arXiv},
      primaryClass={cs.RO}
}

@article{Wang2025_arxiv,
      title={MPPI-DBaS: Safe Trajectory Optimization with Adaptive Exploration}, 
      author={Fanxin Wang and Yikun Cheng and Chuyuan Tao},
      year={2025},
      journal={arxiv:2502.14387},
      archivePrefix={arXiv},
      primaryClass={eess.SY},
}

@inproceedings{Zhang2025SafeVLA,
  title     = {{SafeVLA}: Towards Safety Alignment of Vision-Language-Action Model via Constrained Learning},
  author    = {Zhang, Borong and Zhang, Yuhao and Ji, Jiaming and Lei, Yingshan and Dai, Josef and Chen, Yuanpei and Yang, Yaodong},
  booktitle = {Advances in Neural Information Processing Systems},
  volume    = {38},
  year      = {2025}
}

@article{Hu2025VLSA,
      title={VLSA: Vision-Language-Action Models with Plug-and-Play Safety Constraint Layer}, 
      author={Songqiao Hu and Zeyi Liu and Shuang Liu and Jun Cen and Zihan Meng and Shihefeng Wang and Xiang Li and Xiao He},
      year={2026},
      eprint={2512.11891},
      journal={arXiv},
      primaryClass={cs.RO}
}

@ARTICLE{Zheng2024_TRO,
  author={Zheng, Chunran and Xu, Wei and Zou, Zuhao and Hua, Tong and Yuan, Chongjian and He, Dongjiao and Zhou, Bingyang and Liu, Zheng and Lin, Jiarong and Zhu, Fangcheng and Ren, Yunfan and Wang, Rong and Meng, Fanle and Zhang, Fu},
  journal={IEEE Transactions on Robotics}, 
  title={FAST-LIVO2: Fast, Direct LiDAR–Inertial–Visual Odometry}, 
  year={2025},
  volume={41},
  number={},
  pages={326-346}
  }

@inproceedings{Harabor2011JPS,
  author={Harabor, Daniel and Grastien, Alban},
  title={Online Graph Pruning for Pathfinding on Grid Maps},
  booktitle={Proceedings of the Twenty-Fifth AAAI Conference on Artificial Intelligence},
  volume={25},
  number={1},
  pages={1114--1119},
  year={2011},
  doi={10.1609/aaai.v25i1.7994}
}

@INPROCEEDINGS{Anderson2018,
  author={Anderson, Peter and Wu, Qi and Teney, Damien and Bruce, Jake and Johnson, Mark and Sünderhauf, Niko and Reid, Ian and Gould, Stephen and Van Den Hengel, Anton},
  booktitle={CVPR},
  title={Vision-and-language navigation: Interpreting visually-grounded navigation instructions in real environments},
  year={2018}
}

@INPROCEEDINGS{Krantz2020,
  author={Krantz, Jacob and Wijmans, Erik and Majumdar, Arjun and Batra, Dhruv and Lee, Stefan},
  booktitle={ECCV},
  title={Beyond the nav-graph: Vision-and-language navigation in continuous environments},
  year={2020}
}

@INPROCEEDINGS{Hong2023b,
  author={Hong, Yining and Zhen, Haoyu and Chen, Peihao and Zheng, Shuhong and Du, Yilun and Chen, Zhenfang and Gan, Chuang},
  booktitle={NeurIPS},
  title={3dllm: Injecting the 3d world into large language models},
  year={2023}
}

@ARTICLE{Huang2024c,
  author={Huang, Haifeng and Wang, Zehan and Huang, Rongjie and Liu, Luping and Cheng, Xize and Zhao, Yang and Jin, Tao and Zhao, Zhou},
  journal={arXiv preprint},
  title={Chat-scene: Bridging 3d scene and large language models with object identifiers},
  year={2024}
}

@INPROCEEDINGS{Ding2024,
  author={Ding, Pengxiang and Zhao, Han and Wang, Zhitao and Wei, Zhenyu and Lyu, Shangke and Wang, Donglin},
  booktitle={ECCV},
  title={Quar-vla: Vision-language-action model for quadruped robots},
  year={2024}
}

@inproceedings{Cheng2025,
    archiveprefix = {arxiv},
    eprint        = {2412.04453},
    title={NaVILA: Legged Robot Vision-Language-Action Model for Navigation},
    author={An-Chieh Cheng and Yandong Ji and Zhaojing Yang and Zaitan Gongye and Xueyan Zou and Jan Kautz and Erdem Biyik and Hongxu Yin and Sifei Liu and Xiaolong Wang},
    booktitle   = {Robotics: Science and Systems (RSS)},
    month       = {June},
    year        = {2025},
    website     = {https://navila-bot.github.io},
}

@article{Bu2025,
  title={Univla: Learning to act anywhere with task-centric latent actions},
  author={Bu, Qingwen and Yang, Yanting and Cai, Jisong and Gao, Shenyuan and Ren, Guanghui and Yao, Maoqing and Luo, Ping and Li, Hongyang},
  journal={arXiv preprint arXiv:2505.06111},
  year={2025}
}

@article{Chen2025,
  author = {Xiaoyu Chen and Hangxing Wei and Pushi Zhang and Chuheng Zhang and Kaixin Wang and Yanjiang Guo and Rushuai Yang and Yucen Wang and Xinquan Xiao and Li Zhao and Jianyu Chen and Jiang Bian},
  title = {villa-X: Enhancing Latent Action Modeling in Vision-Language-Action Models},
  journal = {arXiv preprint arXiv:2507.23682},
  year = {2025}
}

@article{Sapkota2025,
  author = {Sapkota, Ranjan and Cao, Yang and Roumeliotis, Konstantinos I. and Karkee, Manoj},
  title = {Vision-Language-Action Models: Concepts, Progress, Applications and Challenges},
  journal = {arXiv preprint arXiv:2505.04769},
  year = {2025}
}
}

\end{document}